%% file: main.tex
\documentclass{article}

\usepackage{microtype}
\usepackage{graphicx}
\usepackage{subfigure}
\usepackage{booktabs} 

\usepackage{hyperref}

\usepackage[accepted]{icml2019}

\input{mathdefinitions}

\icmltitlerunning{Exploration with Unreliable Intrinsic Reward in MARL}

\begin{document}

\twocolumn[
\icmltitle{Exploration with Unreliable Intrinsic Reward \\ 
			in Multi-Agent Reinforcement Learning}



\icmlsetsymbol{equal}{*}

\begin{icmlauthorlist}
\icmlauthor{Wendelin B\"ohmer}{ox}
\icmlauthor{Tabish Rashid}{ox}
\icmlauthor{Shimon Whiteson}{ox}
\end{icmlauthorlist}

\icmlaffiliation{ox}{Department of Computer Science, University of Oxford, United Kingdom}

\icmlcorrespondingauthor{Wendelin B\"ohmer}{wendelin.boehmer@cs.ox.ac.uk}

\icmlkeywords{Exploration for RL Workshop, ICML, Exploration, MARL, IQL, SMAC, Intrinsic Reward, ICQL}

\vskip 0.3in
]



\printAffiliationsAndNotice{}  

\setlength{\abovedisplayskip}{1.5mm}
\setlength{\belowdisplayskip}{1.5mm}

\begin{abstract}
This paper investigates the use of intrinsic reward
to guide exploration in multi-agent reinforcement learning.
We discuss the challenges in applying intrinsic reward  
to multiple collaborative agents
and demonstrate how unreliable reward can 
prevent decentralized agents from learning the optimal policy.
We address this problem with a novel framework,
Independent Centrally-assisted Q-learning (ICQL),
in which decentralized agents share control and 
an experience replay buffer with a centralized agent.
Only the centralized agent is intrinsically rewarded,
but the decentralized agents still benefit from improved exploration,
without the distraction of unreliable incentives. 
\end{abstract}

\input{introduction}

\input{background}
\input{method}

\input{experiments}

\section*{Acknowledgements}
The authors would like to thank 
Jakob F\"orster, Gregory Farquhar
and Christian Schroeder de Witt
for fruitful discussions
about decentralization and exploration in MARL. 
This project has received funding 
from the European Research Council (ERC), 
under the European Union's Horizon 2020 research 
and innovation programme (grant agreement number 637713),
and a grant of the EPSRC (EP/M508111/1, EP/N509711/1).
The experiments were made possible 
by a generous equipment grant from NVIDIA.

\bibliography{bibliography}
\bibliographystyle{icml2019}
\newpage
\part*{Appendix}
\appendix
\input{appendix}
\input{relatedwork}

\end{document}

%% file: mathdefinitions.tex
\usepackage{amsmath,amssymb,amsfonts,bm}

\def\R{\mathrm{I\kern-0.4ex R}}
\def\N{\mathrm{I\kern-0.4ex N}}
\def\E{\mathrm{I\kern-0.4ex E}}
\newcommand{\Set}[1]{\mathcal{#1}}

\newcommand{\mat}[1]{\mathbf{#1}}
\newcommand{\ve}[1]{\bm #1}

\DeclareMathOperator*{\localmax}{lmax}

\newcommand{\smallsum}[2]{ {\textstyle 
	\sum\limits_{\scriptscriptstyle #1}^{\scriptscriptstyle #2}} 
}

%% file: introduction.tex
\section{Introduction}
Recent successes in challenging computer games 
like StarCraft 2 \citep{Vinyals19}
have raised interest in Multi-agent Reinforcement Learning (MARL).
Here single units are modeled as individual agents,
for example, in the recent open source StarCraft Multi-agent Challenge 
\citep[SMAC,][]{Samvelyan19}. 
In comparison to single-agent deep RL,
MARL faces some unique challenges,
in particular {\em decentralization} and {\em coordination}. 
In this paper we investigate the equally challenging problem of {\em directed exploration}.

Directed exploration in single-agent deep RL 
still poses many open questions,
like how to generalize \emph{visitation counts} 
in large input spaces and
how to change the exploration policy quickly towards
newly discovered states.
However, so far,
there has been little work on exploration for deep MARL.
Exploration in MARL differs from the single-agent 
setting in some challenging ways: 
(i) counting visitations of state-action pairs 
is infeasible for many agents,
due to the large joint-action space;
(ii) as unexpected outcomes can be caused by multiple agents, 
it is not clear which agent's action should be reinforced; and 
(iii) partial observability decreases 
the reliability of count estimates. 

Decentralization is required in MARL when the agents 
cannot communicate directly. 
Moreover, a centralized control policy is often infeasible,
as the joint-action space grows exponentially in the number of agents.
In line with SMAC, we consider the case where the state of the system 
is only partially observable by each agent,
although during training the global state may be available. 
This is called \emph{centralized training 
with decentralized execution} \citep{Foerster16}.
This paper focuses on the  
simplest value-based algorithm in this class, 
a variant of Independent Q-learning \citep[IQL,][]{Tan93},
where each agent acts on partial observations 
and assumes the other agents' decisions
are an unobserved, stationary part of the environment. 

However, these simple decentralized agents  lack coordination.
Take the example of two predators,  
who trapped their prey in a corner. 
To catch it, both must attack simultaneously, 
as it will escape if only one predator attacks.
From the perspective of each predator, 
the reward for attacking depends on the actions of the other. 
When the punishment for letting the
prey escape is larger than the reward for catching it, 
neither agent will learn the optimal strategy independently.
There are multiple methods using centralized training
to mitigate this effect for decentralized policies,
e.g., multi-agent credit assignment \citep[COMA,][]{Foerster17}
and bootstrapping with an approximation of
the joint Q-value function. 
For example, Value Decomposition Networks \citep[VDN,][]{Sunehag17}
optimize the joint Q-value, 
but restrict the Q-value function 
to a sum of individual agents' utilities. 
QMIX \citep{Rashid18} goes one step further 
and mixes the agents' utilities with a non-linear hyper-network, 
that conditions on the global state. 
Both approaches can execute the decentralized learned policy  
by maximizing each agent's utility.
However, all the above techniques use 
relatively simple $\epsilon$-greedy exploration.

Value-based algorithms that explore provably efficiently
in a tabular setting \citep[e.g.][]{Jin18}
rely on \emph{optimism in the face of uncertainty}.
There are two major lines of research in the literature:
to use {\em intrinsic reward} to over-estimate uncertain state-action values
or to use Thompson sampling from a 
{\em Bayesian posterior} of the value function.
Various techniques have been proposed to estimate the 
uncertainty of state-action values 
(summarized in Appendix \ref{sec:related}), 
but whether it is used as an intrinsic reward
or as the standard deviation of a Gaussian posterior,
most works converge at an estimate 
that is supposed to scale with $1/\sqrt{N_t(s_t, u_t)}$,
where $N_t(s_t,u_t)$ counts how often state $s_t$ and action $u_t$ 
have been observed at time $t$.
For large input spaces, however,
these estimates are rough approximations of visitation counts
and the resulting uncertainties are highly unreliable.

This paper investigates estimated intrinsic reward 
for decentralized IQL agents.
We evaluate the {\em variance of linear functions} \citep{ODonoghue18}
as an uncertainty estimate 
in a novel predator-prey task that emphasizes exploration.
We observe empirically that the intrinsic reward accelerates learning, 
but remains inherently {\em unreliable}, 
which prevents finding the optimal solution.
To learn reliable decentralized policies in the face of unreliable reward,
we propose to share control with a second agent 
that is discarded after training and can thus be centralized.
Only the central agent receives intrinsic rewards, 
which prevents the decentralized agents from being distracted, 
while still improving their exploratory behavior.
We show that this new approach to MARL exploration 
drastically speeds up learning of the decentralized agents, 
while converging to the optimal solution. 
This novel framework is general 
and can be applied to different estimators of 
intrinsic reward and/or off-policy MARL algorithms like VDN and QMIX.

%% file: background.tex
\section{Background}
We restrict ourselves to {\em cooperative tasks},
modeled as a Dec-POMDP \citep{Oliehoek16}, that is, 
a tuple $\langle \Set S, \{\Set U^a\}, P, r, \{\Set Z^a\}, 
\{O^a\}, n, \gamma \rangle$.
The global state of the system is denoted as $s \in \Set S$,
and each of the $n$ agents chooses actions $u^a \in \Set U^a$,
which together form the joint action $\ve u \in \Set U$.
After executing joint action $\ve u_t$ in state $s_t$ 
at discrete time step $t$, 
the next state $s_{t+1}$ 
is drawn from transition kernel $P(s_{t+1}|s_t, \ve u_t)$, 
and a reward $r_t := r(s_t, \ve u_t)$ 
is determined by the reward function 
$r: \Set S \times \Set U \to \R$.
While a {\em centralized joint policy} $\pi_c(\ve u | s_t)$
can choose joint actions $\ve u$ conditioned 
on the current state $s_t$,
a {\em decentralized agent policy} $\pi^a(u^a|\tau^a_t)$ 
draws only agent $a$'s action $u^a \in \Set U^a$,
based on the agent's current trajectory $\tau^a_t$ 
of past actions $u^a_i$ and observations $z^a_i \in \Set Z^a$,
which are drawn from the agent's observation kernel 
$O^a(z^a_i|s_i)$, that is,
$\tau^a_t := [z^a_0, u^a_0, z^a_1, u^a_1, \ldots, z^a_t]$.
Execution of a joint policy $\pi$ yields an episode
with return $R_t := \sum_{i=t}^\infty \gamma^{i-t} r_i$
at time $t$.
Our goal is to find a decentralized joint policy
$\pi(\ve u|\{\tau^a_t\}) := \prod_{a=1}^n \pi^a(u^a|\tau^a_t)$,
which maximizes the expected return for each observed trajectory. 
Partial observability of the policy
can significantly slow down learning 
and we allow access to the global state during training,
that is, centralized training for decentralized execution.

\subsection{Independent Q-learning (IQL)} \label{sec:bg_iql}
Independent Q-learning \citep{Tan93}
approaches this goal by defining the state-action value function
of agent $a$ as the expectation of the return, 
following policy $\pi$ from an observed trajectory $\tau^a_t$, that is,  
$Q^a(u^a|\tau^a_t) := \E_\pi[R_t|\tau^a_t, u^a]$. 
As in Q-learning \citep{Watkins92},
IQL assumes that the greedy policy, 
which chooses the action with the largest corresponding value,
maximizes the expected return in each state.
Note that this is not true,
as the expected return also depends on the other agents' behavior,
which can introduce non-stationarity.
That being said, IQL appears to be stable in practice
and works quite well in most tasks.

We use a neural network with one head for each discrete action
to approximate the value function \citep[as in DQN,][]{Mnih15}.
For IQL, we learn a function\footnote{To condition on trajectories $\tau^a_t$, 
	we follow \citet{Hausknecht15}
	and use a recurrent network of GRU units \cite{Chung14}, 
	which condition on their hidden state, the last action $u^a_{t-1}$,
	the current observation $z^a_t$ and the agent id $a$. 
} $q^a_\theta(u^a|\tau^a_t)$,
parameterized by $\theta$, 
with gradient-descend on the expected squared Bellman error
\begin{equation}
	\min_\theta \; \E\Big[\smallsum{t=0}{\infty} \big(
		r_t + \gamma \max_{u'} q^a_{\theta'}(u'|\tau^a_{t+1}) 
		- q^a_\theta(u^a_t|\tau^a_t)
	\big)^2 \Big] \,,
\end{equation}
where $\theta'$ are the parameters of a target network,
which are replaced with a copy of the current parameters $\theta$
from time to time to improve stability.
The expectation is approximated by drawing 
mini-batches of trajectories from an experience replay buffer
\citep{Lin92}.
We also use double Q-learning \citep{vanHasselt16} 
to further improve stability
and share the parameters $\theta$ of all agents' value functions
for better generalization \citep[similar to QMIX,][]{Rashid18}.

\subsection{Intrinsic Reward} \label{sec:bg_ir}
We employ a local uncertainty measure introduced by \citet{ODonoghue18}.
The variance of a linear regression,
i.e., fitting a linear function $f(u|s) = \ve w_u^\top \ve\phi(s)$
to a fixed set of state-action pairs $\{s_i, u_i\}_{i=1}^t$
and random labels $y_i$ with a Gaussian distribution 
$\Set N(y_i|y(s_i), \sigma^2)$, is
\begin{equation} \label{eq:linvar}
	V_t[f](u|s) \;=\; \sigma^2 \; \ve\phi(s)^{\!\top}\! 
		\Big(\smallsum{i=1}{t} \delta_{u_i u} \, 
			\ve\phi(s_i) \ve\phi(s_i)  
		\Big)^{\!-1} \!\!\!\!\ve\phi(s) \,.
\end{equation}
As each head of the IQL value function $q^a(u^a|\tau^a_t)$
is a linear function of the last network layer, 
that is, the hidden state $\ve\phi^a(\tau^a_t)$ of a GRU,
\citet{ODonoghue18} suggest to use $r^+_t := \sqrt{V_t[q^a](u^a_t|\tau^a_t)}$
as a measure of local uncertainty. 
This choice of intrinsic reward is somewhat justified,
as for one-hot coded $\ve\phi(s_t)$, 
the intrinsic reward is $r^+_t = \sigma / \sqrt{N_t(s_t, u_t)}$,
which corresponds to the tabular case 
\citep[e.g., in][]{Jin18} with scaling factor $\sigma$.  

%% file: method.tex
\section{Method}
Intrinsic reward based on estimated uncertainties 
rarely reflects the precise visitation counts.
We investigate how such unreliable intrinsic reward
can still reliably improve exploration of decentralized agents.
The main idea is to introduce a second controller during training that can be discarded afterwards.
This joint agent is intrinsically rewarded and 
can thus explore the environment in a directed fashion.
In principle, the agent's policy could be learned 
by many algorithms.
As it is only active during training, though,
we propose a {\em centralized agent},
which conditions on the more informative global state. 
Most importantly,  
we train simultaneously the {\em decentralized agents}, 
which will be later deployed for execution,
on the same replay buffer.
These can utilize any decentralized off-policy learning algorithm,
but we focus in the following on IQL for simplicity.

\input{figure1}

\subsection{A Central MARL Agent}
The large action spaces in MARL make individual heads for each 
joint action $\ve u$ on value functions 
infeasible in the face of many agents.
Maximizing a value function that conditions on all agents' actions,
on the other hand,
has to be evaluated for all $\ve u$ as well,
which can be prohibitively expensive.
Instead, we use the architecture of a COMA critic,
which \citet{Foerster17} introduced 
in the context of a policy gradient method.
They define an agent-specific joint-value function $q^a_\psi$,
parameterized by $\psi$, 
which has a head for each of $a$'s actions $u^a_t$ 
and conditions on the global state, all other agents' actions 
$\ve u^{-a}_t := [u^1_t, \ldots, u^{a-1}_t, u^{a+1}_t, \ldots, u^n_t]$
and agent $a$'s the previous action $u^a_{t-1}$:
\vspace{-1mm}
\begin{equation}
	q^a_\psi(u^a_t|s_t, \ve u^{-a}_t, u^a_{t-1}) 
	\quad \stackrel{!}{\approx} \quad \E[R_t | s_t, \ve u_t] \,.
\end{equation}
Instead of maximizing this function
w.r.t.~the joint action $\ve u_t$,
we propose here to approximate a local maximum by iteratively 
choosing the $u^a_t$ that maximizes each individual agent's $q^a_\psi$,
and using it for $\ve u^{-a}$ in the next iteration of the maximization. 
In practice, we initialize $\ve u_t$ with the greedy actions of the 
decentralized IQL agents and then perform this iterative 
{\em local maximization} (denoted $\localmax$) 
for a small number of iterations.
As in IQL, agents share parameters $\psi$.

During exploration it is important that the sampling 
policy changes in response to newly discovered states.
This change is imposed by intrinsic reward,
which must therefore be transported quickly 
to earlier states in the episode.
We use a $Q(\lambda)$ implementation \citep{Watkins89} 
to accelerate this transport,
but do not cut the traces after exploratory steps.
This improves transport, 
but also introduces non-stationary targets.
Our training procedure performs a parameter update
after each episode and we compute the targets backwards:
\begin{equation} \nonumber
	G_t^\lambda := r_t + (1-\lambda) \gamma 
		\localmax\limits_{\bar{\ve u}}  
		q^a_{\psi'}(\bar u^a|s_{t+1}, \ve \bar{\ve u}^{-a}, u^a_{t})
		+ \lambda \gamma G^\lambda_{t+1} \,,
\vspace{-1mm}
\end{equation}
where $\psi'$ denotes the target network
and $G_{T}^\lambda := 0$.
The loss
\vspace{-4mm}
\begin{equation} 
	\min\limits_\psi \; \E \Big[\smallsum{t=0}{T-1} \smallsum{a=1}{n}
	\big( G_t^\lambda		
		- \, q^a_\psi(u^a_t|s_t, \ve u^{-a}_t, u^a_{t-1}) \big)^2 \Big] \,,
\end{equation}
is minimized by gradient descent,
where the expectation is approximated with the
same mini-batches as in IQL 
and the same stabilization techniques are used as well.

\subsection{Intrinsic Reward Revisited}
\label{sec:ir}
Intrinsic rewards, as defined in Section \ref{sec:bg_ir},
induce three challenges for collaborative MARL:
dependence on joint actions, collaborative rewards, 
and evolving parameters. 

First, 
estimating \eqref{eq:linvar} is infeasible for MARL,
as we would have to estimate one correlation matrix 
for each joint action $\ve u$.
Instead of estimating a measure that depends 
on counting $N_t(s_t,\ve u_t)$, 
we propose here to estimate one based on the count $N_t(s_{t+1})$.
Although only a heuristic, this approach works in arbitrary large action spaces.

Second, in collaborative tasks all agents 
should receive the same reward to avoid diverging incentives.
However, in MARL each agent $a$ estimates a different uncertainty,
based on $a$'s observations and/or other inputs.
As the interaction of all agents could have contributed 
to each agent's uncertainty, 
it is unclear how to reconcile diverging estimates. 
We propose to use the largest uncertainty as intrinsic reward
for all agents, to consider this potential interaction.

Third, 
the agents' value function parameters continually change, 
particularly in the beginning of training
when exploration is most important. 
The same inputs $x$ yield different 
representations $\ve\phi^a(x)$ at different times $t$,
and estimates with \eqref{eq:linvar} therefore 
become outdated after a while. 
To reflect this change, 
we propose to use an exponentially decaying average
of the inverted matrix
$\mat C_t := (1-\alpha) \, \mat C_{t-1} 
+ \sum_{a=1}^n \ve \phi^a_{(x_{t})} \, \ve \phi^{a\top}_{(x_{t})}$,
where $x_{t}$ denotes the value function's inputs at time $t$
and $\alpha$ is a small decay constant.
As the resulting uncertainty never decays to 0,
we also introduce a {\em bias} $b_t$ and discard negative intrinsic rewards.
The bias can be constant or an average over past uncertainties
to only reward states with above average novelty. 

The resulting collaborative intrinsic reward $r^+_t$ is:
\begin{equation}  
	r^+_t := \sigma \max\!\Big\{ 0, \max_{a}\!\Big(
		\sqrt{\ve\phi^{a\top}_{(x_{t+1})} \mat C_t^{-1} 
		\ve\phi^a_{(x_{t+1})}} - b_t \Big) 
	\Big\} . \!\!
\end{equation}

\subsection{Independent Centrally-assisted Q-learning (ICQL)}
\label{sec:icql}
The intrinsically rewarded central agent
samples in our training framework episodes, 
while the decentralized (here IQL) agents are simultaneously trained
on the shared replay buffer.
This improves exploration in two ways:
(i) although the decentralized agents still 
benefit from the exploration that is 
induced by (potentially unreliable) intrinsic reward,
their final policies are exclusively based on environmental rewards, and
(ii) the central agent conditions on the true state of the system,
which includes information 
that may not be observable by the agents.

However, sampling only with the central agent
yields an out-of-distribution problem 
for the IQL agents:
a single deviating decision can induce 
a trajectory that has never been seen during training. 
We therefore share the sampling process
by deciding randomly at the start of an episode
whether the IQL agents or the central agent takes control.
The resulting behavior appears quite stable
for different probabilities of that choice
and we chose 50\% for simplicity.

%% file: figure1.tex
\begin{figure*}[t!]
	\begin{center}
		\includegraphics[width=0.95\textwidth]{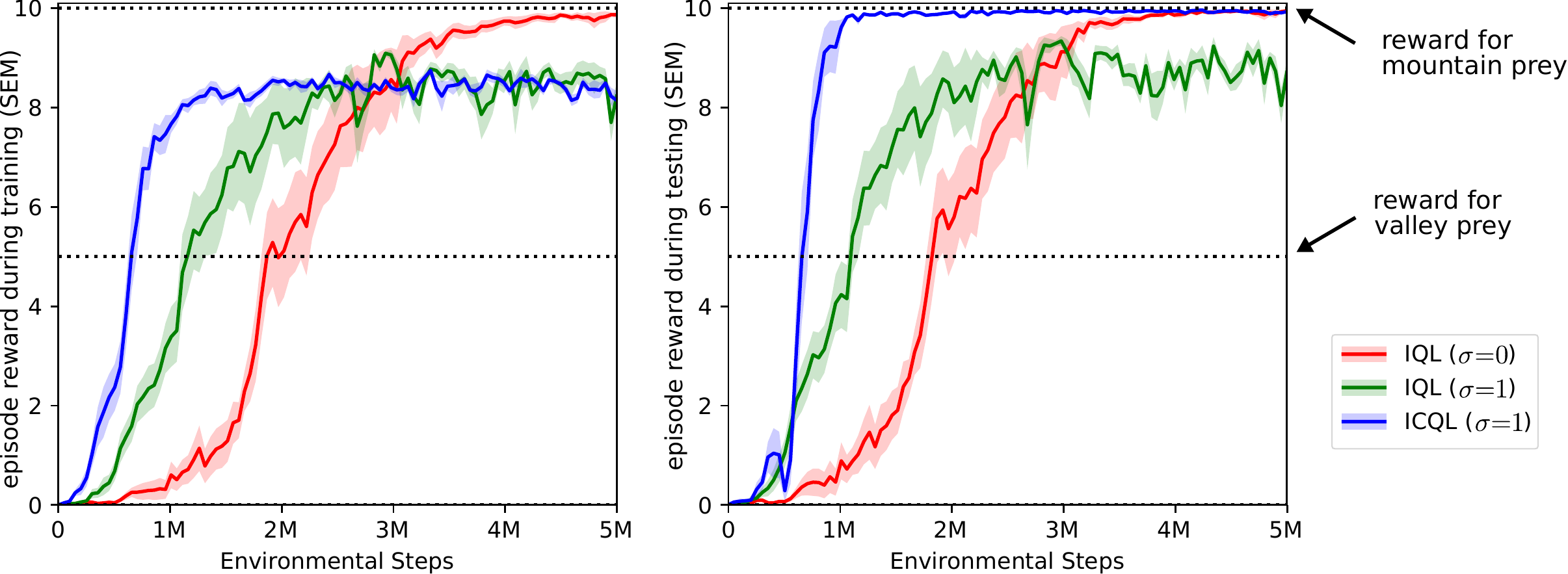}
	\end{center}
	\vspace{-5mm}
	\caption{ \label{fig:mountain}
			4 agents hunt a mountain and a valley prey.
			We plot mean and standard error (over 8 seeds) 
			of the training (left) and test performance (right) of 
			IQL, with and without {\em intrinsic reward} 
			(i.e.~magnitude $\sigma$),
			and of our centrally-assisted exploration framework ICQL.
			Note that for ICQL training performance is 50\% centralized,
			but test performance is 100\% decentralized.}
	\vspace{-1mm}
\end{figure*}

%% file: experiments.tex
\section{Experiments}
We extend a common partially observable predator-prey task in a grid-world
to make it challenging from an exploration perspective,
as preliminary experiments have shown that the 
original task does not require directed 
exploration of the state space.\footnote{In the original task, the prey moves randomly 
	and the states in which the agents meet it
	are almost uniformly distributed.
	This provides sufficient exploration to find an optimal policy
	and directed exploration is unnecessary.
} We train 4 agents, 
with $5 \times 5$ agent-centric observations,
to hunt a mountain and a valley prey.
Prey moves randomly in a bounded grid-world 
of height 41 and width 10. 
To simulate a {\em mountain}, 
both agents and valley prey do not execute 
50\% of all `up' actions, 
and mountain prey does not execute 
50\% of all `down' actions.
Valley prey spawn randomly on the lowest row,
mountain prey on the highest row, and 
agents on middle row.
An episode ends either when one of the prey is caught, 
that is, when agents/boundaries surround it on all sides,
or after 100 steps.
Only capturing yields reward,
5 for the valley and 10 for the mountain prey.
Exploring the state space helps to find the
mountain prey without getting distracted by the valley prey.

Figure \ref{fig:mountain} shows training and test performance\footnote{We implemented all algorithms in the 
	PyMarl framework \citep{Samvelyan19}.
	Details can be found in Appendix \ref{sec:details}.
}
for three algorithms:
the original IQL (red, Section \ref{sec:bg_iql}),
IQL with intrinsic reward based on the agents' last layers
(green, Section \ref{sec:ir}) and our novel
centrally-assisted exploration framework ICQL
(blue, Section \ref{sec:icql}),
where the intrinsic reward is based on the last layer 
of the central agents' value functions.

The destabilizing effect of unreliable intrinsic reward 
on IQL can be seen in the green IQL ($\sigma$=1) curve:
it speeds up learning, 
but also prevents the agents from finding the optimal policy
(visible both in training and test plots).
The bonus provides incentives for exploration,
but also appears to distract the agents
when their policy should converge. 

Our ICQL framework (blue) learns even faster,
but demonstrates different behavior during training and testing.
On the one hand, one can see the same sub-optimal behavior 
during training (left plot), which executes 50\% of the
episodes with the intrinsically-rewarded central agent 
and 50\% with decentralized agents trained simultaneously.
On the other hand, the test performance (right plot)
of the decentralized agents
shows the same improved learning, 
but none of the instabilities once the mountain prey has been found. 

We conclude that
intrinsic reward is both a blessing and a curse for MARL settings.
We have shown that even unreliable reward can improve directed exploration,
but also introduces detracting incentives.
Our novel ICQL framework for centrally-assisted exploration
appears to stabilize learning
and further speeds up training,
most likely by exploiting access to the true state.

In future work, we will further evaluate how the framework 
performs with different decentralized learning algorithms, 
like VDN and QMIX,
and employ other uncertainty estimates for intrinsic rewards. 
We also want to investigate the effect of adaptive biases
and apply our method to StarCraft micromanagement tasks 
\citep[SMAC,][]{Samvelyan19}.

%% file: appendix.tex
\section{Training details} \label{sec:details}
	We implemented all algorithms in the PyMARL framework \citep{Samvelyan19}, 
	where we used RMSprop with learning rate 0.0005,
	$\gamma=0.99$, batch size 32 
	and a replay buffer holding the last 200 episodes.
	Decentralized agents had a hidden layer of 64 GRU cells,
	sandwiched between 2 feed-forward layers,
	and central agents had 3 feed-forward layers with 128 hidden neurons each.
	The target network was updated every 200 episodes 
	and we used $\epsilon$-greedy exploration, 
	which decayed $1 \geq \epsilon \geq 0.05$ within 20,000 steps.
	Intrinsic reward had magnitude $\sigma=1$, 
	a decay constant $\alpha=0.0002$ and constant bias $b_t:=0.01$.
	ICQL approximated the local maximum with one $\localmax$ iteration. 

%% file: relatedwork.tex
\section{Related Work}
\label{sec:related}

In this paper we focus on intrinsic reward for exploration
\citep[in difference to pure curiosity,][]{Burda19b}.
Here the uncertainty is often derived from to the 
prediction quality after training on past trajectories.
For example, pseudo-counts are based on the reconstruction probability of 
visual observations \citep{Bellemare2016,Ostrovski2017},
using a PixelCNN \citep{vandenOord16}.
Alternatively, \citet{Tang17} count visitations
using a hash function on a random linear projection.
Furthermore, \citet{Pathak17} use the predictability 
of the observed transition as intrinsic reward signal,
and \citet{Roderick18} reduce uncertainty 
with prior knowledge over state abstractions.

In the context of Bayesian posteriors,
the uncertainty has been estimated from
an ensemble of value functions, 
with optional bootstrapping techniques
\citep{Osband16a, Osband18}. 
Alternatively, Noisy Nets \citep{Fortunato18,Plappert18}
sample a value function for each episode from 
a diagonal Gaussian posterior over the parameters of the neural network. 
Similarly, \citet{Gal17} suggested to use Concrete Dropout 
to estimate the posterior for model-based RL.
To include the uncertainty of future reward,
\citet{ODonoghue18} proposed the 
Uncertainty Bellman Equation (UBE), 
which propagates the `local uncertainty' of future decisions 
with a Bellman operator.

For MARL,
\citet{Zheng18} proposed to coordinate exploration 
by sharing latent variables, drawn from a learned distribution. 
\citet{Jaques18} focuses on social motivations of competitive agents
and \citet{Leibo19} describes exploration as an auto-curriculum  
generated by competing species of agents.

%% file: main.bbl
\begin{thebibliography}{32}
\providecommand{\natexlab}[1]{#1}
\providecommand{\url}[1]{\texttt{#1}}
\expandafter\ifx\csname urlstyle\endcsname\relax
  \providecommand{\doi}[1]{doi: #1}\else
  \providecommand{\doi}{doi: \begingroup \urlstyle{rm}\Url}\fi

\bibitem[Bellemare et~al.(2016)Bellemare, Srinivasan, Ostrovski, Schaul,
  Saxton, and Munos]{Bellemare2016}
Bellemare, M.~G., Srinivasan, S., Ostrovski, G., Schaul, T., Saxton, D., and
  Munos, R.
\newblock Unifying count-based exploration and intrinsic motivation.
\newblock In \emph{Advances in Neural Information Processing Systems (NIPS)
  29}, pp.\  1471--1479, 2016.

\bibitem[Burda et~al.(2019)Burda, Edwards, Pathak, Storkey, Darrell, and
  Efros]{Burda19b}
Burda, Y., Edwards, H., Pathak, D., Storkey, A., Darrell, T., and Efros, A.~A.
\newblock Large-scale study of curiosity-driven learning.
\newblock In \emph{International Conference on Learning Representations
  (ICLR)}, 2019.

\bibitem[Chung et~al.(2014)Chung, Gulcehre, Cho, and Bengio]{Chung14}
Chung, J., Gulcehre, C., Cho, K., and Bengio, Y.
\newblock Empirical evaluation of gated recurrent neural networks on sequence
  modeling.
\newblock In \emph{NIPS Workshop on Deep Learning}, 2014.
\newblock URL \url{http://arxiv.org/abs/1412.3555}.

\bibitem[Foerster et~al.(2016)Foerster, Assael, de~Freitas, and
  Whiteson]{Foerster16}
Foerster, J., Assael, I.~A., de~Freitas, N., and Whiteson, S.
\newblock Learning to communicate with deep multi-agent reinforcement learning.
\newblock In \emph{Advances in Neural Information Processing Systems (NIPS)
  29}, pp.\  2137--2145. 2016.

\bibitem[Foerster et~al.(2018)Foerster, Farquhar, Afouras, Nardelli, and
  Whiteson]{Foerster17}
Foerster, J.~N., Farquhar, G., Afouras, T., Nardelli, N., and Whiteson, S.
\newblock Counterfactual multi-agent policy gradients.
\newblock In \emph{Proceedings of the 15th {AAAI} Conference on Artificial
  Intelligence}, pp.\  2974--2982, 2018.

\bibitem[Fortunato et~al.(2018)Fortunato, Azar, Piot, Menick, Hessel, Osband,
  Graves, Mnih, Munos, Hassabis, Pietquin, Blundell, and Legg]{Fortunato18}
Fortunato, M., Azar, M.~G., Piot, B., Menick, J., Hessel, M., Osband, I.,
  Graves, A., Mnih, V., Munos, R., Hassabis, D., Pietquin, O., Blundell, C.,
  and Legg, S.
\newblock Noisy networks for exploration.
\newblock In \emph{International Conference on Learning Representations
  (ICLR)}, 2018.

\bibitem[Gal et~al.(2017)Gal, Hron, and Kendall]{Gal17}
Gal, Y., Hron, J., and Kendall, A.
\newblock Concrete dropout.
\newblock In \emph{Advances in Neural Information Processing Systems (NIPS)},
  pp.\  3584--3593, 2017.

\bibitem[Hasselt et~al.(2016)Hasselt, Guez, and Silver]{vanHasselt16}
Hasselt, H.~v., Guez, A., and Silver, D.
\newblock Deep reinforcement learning with double q-learning.
\newblock In \emph{Proceedings of the 13th AAAI Conference on Artificial
  Intelligence}, pp.\  2094--2100, 2016.

\bibitem[Hausknecht \& Stone(2015)Hausknecht and Stone]{Hausknecht15}
Hausknecht, M.~J. and Stone, P.
\newblock Deep recurrent q-learning for partially observable mdps.
\newblock In \emph{2015 {AAAI} Fall Symposia}, pp.\  29--37, 2015.
\newblock URL
  \url{http://www.aaai.org/ocs/index.php/FSS/FSS15/paper/view/11673}.

\bibitem[Jaques et~al.(2018)Jaques, Lazaridou, Hughes, G{\"{u}}l{\c{c}}ehre,
  Ortega, Strouse, Leibo, and de~Freitas]{Jaques18}
Jaques, N., Lazaridou, A., Hughes, E., G{\"{u}}l{\c{c}}ehre, {\c{C}}., Ortega,
  P.~A., Strouse, D., Leibo, J.~Z., and de~Freitas, N.
\newblock Intrinsic social motivation via causal influence in multi-agent {RL}.
\newblock \emph{CoRR}, abs/1810.08647, 2018.
\newblock URL \url{https://arxiv.org/abs/1810.08647}.

\bibitem[Jin et~al.(2018)Jin, Allen-Zhu, Bubeck, and Jordan]{Jin18}
Jin, C., Allen-Zhu, Z., Bubeck, S., and Jordan, M.~I.
\newblock Is {Q}-learning provably efficient?
\newblock In \emph{Advances in Neural Information Processing Systems (NeurIPS)
  31}, pp.\  4863--4873. 2018.

\bibitem[Leibo et~al.(2019)Leibo, Hughes, Lanctot, and Graepel]{Leibo19}
Leibo, J.~Z., Hughes, E., Lanctot, M., and Graepel, T.
\newblock Autocurricula and the emergence of innovation from social
  interaction: {A} manifesto for multi-agent intelligence research.
\newblock \emph{CoRR}, abs/1903.00742, 2019.
\newblock URL \url{http://arxiv.org/abs/1903.00742}.

\bibitem[Lin(1992)]{Lin92}
Lin, L.-J.
\newblock Self-improving reactive agents based on reinforcement learning,
  planning and teaching.
\newblock \emph{Machine Learning}, 8\penalty0 (3):\penalty0 293--321, 1992.

\bibitem[Mnih et~al.(2015)Mnih, Kavukcuoglu, Silver, Rusu, Veness, Bellemare,
  Graves, Riedmiller, Fidjeland, Ostrovski, Petersen, Beattie, Sadik,
  Antonoglou, King, Kumaran, Wierstra, Legg, and Hassabis]{Mnih15}
Mnih, V., Kavukcuoglu, K., Silver, D., Rusu, A.~A., Veness, J., Bellemare,
  M.~G., Graves, A., Riedmiller, M., Fidjeland, A.~K., Ostrovski, G., Petersen,
  S., Beattie, C., Sadik, A., Antonoglou, I., King, H., Kumaran, D., Wierstra,
  D., Legg, S., and Hassabis, D.
\newblock Human-level control through deep reinforcement learning.
\newblock \emph{Nature}, 518\penalty0 (7540):\penalty0 529--533, February 2015.

\bibitem[O'Donoghue et~al.(2018)O'Donoghue, Osband, Munos, and
  Mnih]{ODonoghue18}
O'Donoghue, B., Osband, I., Munos, R., and Mnih, V.
\newblock The uncertainty {B}ellman equation and exploration.
\newblock In \emph{Proceedings of the 35th International Conference on Machine
  Learning (ICML)}, pp.\  3836--3845, 2018.

\bibitem[Oliehoek \& Amato(2016)Oliehoek and Amato]{Oliehoek16}
Oliehoek, F.~A. and Amato, C.
\newblock \emph{A concise introduction to decentralized POMDPs}.
\newblock Springer Publishing Company, Incorporated, 1st edition, 2016.
\newblock ISBN 3319289276, 9783319289274.

\bibitem[Osband et~al.(2016)Osband, Van~Roy, and Wen]{Osband16a}
Osband, I., Van~Roy, B., and Wen, Z.
\newblock Generalization and exploration via randomized value functions.
\newblock In \emph{Proceedings of the 33rd International Conference on
  International Conference on Machine Learning (ICML)}, pp.\  2377--2386, 2016.

\bibitem[Osband et~al.(2018)Osband, Aslanides, and Cassirer]{Osband18}
Osband, I., Aslanides, J., and Cassirer, A.
\newblock Randomized prior functions for deep reinforcement learning.
\newblock In \emph{Advances in Neural Information Processing Systems (NeurIPS)
  31}, pp.\  8617--8629. 2018.

\bibitem[Ostrovski et~al.(2017)Ostrovski, Bellemare, van~den Oord, and
  Munos]{Ostrovski2017}
Ostrovski, G., Bellemare, M.~G., van~den Oord, A., and Munos, R.
\newblock Count-based exploration with neural density models.
\newblock In \emph{Proceedings of the 34th International Conference on Machine
  Learning (ICML)}, pp.\  2721--2730, 2017.

\bibitem[Pathak et~al.(2017)Pathak, Agrawal, Efros, and Darrell]{Pathak17}
Pathak, D., Agrawal, P., Efros, A.~A., and Darrell, T.
\newblock Curiosity-driven exploration by self-supervised prediction.
\newblock In \emph{Proceedings of the 34th International Conference on Machine
  Learning (ICML)}, 2017.

\bibitem[Plappert et~al.(2018)Plappert, Houthooft, Dhariwal, Sidor, Chen, Chen,
  Asfour, Abbeel, and Andrychowicz]{Plappert18}
Plappert, M., Houthooft, R., Dhariwal, P., Sidor, S., Chen, R.~Y., Chen, X.,
  Asfour, T., Abbeel, P., and Andrychowicz, M.
\newblock Parameter space noise for exploration.
\newblock In \emph{International Conference on Learning Representations
  (ICLR)}, 2018.

\bibitem[Rashid et~al.(2018)Rashid, Samvelyan, de~Witt, Farquhar, Foerster, and
  Whiteson]{Rashid18}
Rashid, T., Samvelyan, M., de~Witt, C.~S., Farquhar, G., Foerster, J.~N., and
  Whiteson, S.
\newblock {QMIX:} monotonic value function factorisation for deep multi-agent
  reinforcement learning.
\newblock In \emph{International Conference on Machine Learning (ICML)}, pp.\
  4292--4301, 2018.

\bibitem[Roderick et~al.(2018)Roderick, Grimm, and Tellex]{Roderick18}
Roderick, M., Grimm, C., and Tellex, S.
\newblock Deep abstract {Q}-networks.
\newblock In \emph{Proceedings of the 17th International Conference on
  Autonomous Agents and MultiAgent Systems (AAMAS)}, pp.\  131--138, 2018.

\bibitem[Samvelyan et~al.(2019)Samvelyan, Rashid, de~Witt, Farquhar, Nardelli,
  Rudner, Hung, Torr, Foerster, and Whiteson]{Samvelyan19}
Samvelyan, M., Rashid, T., de~Witt, C.~S., Farquhar, G., Nardelli, N., Rudner,
  T. G.~J., Hung, C., Torr, P. H.~S., Foerster, J.~N., and Whiteson, S.
\newblock The {S}tar{C}raft multi-agent challenge.
\newblock \emph{CoRR}, abs/1902.04043, 2019.
\newblock URL \url{https://arxiv.org/abs/1902.04043}.

\bibitem[Sunehag et~al.(2018)Sunehag, Lever, Gruslys, Czarnecki, Zambaldi,
  Jaderberg, Lanctot, Sonnerat, Leibo, Tuyls, and Graepel]{Sunehag17}
Sunehag, P., Lever, G., Gruslys, A., Czarnecki, W.~M., Zambaldi, V., Jaderberg,
  M., Lanctot, M., Sonnerat, N., Leibo, J.~Z., Tuyls, K., and Graepel, T.
\newblock Value-decomposition networks for cooperative multi-agent learning
  based on team reward.
\newblock In \emph{Proceedings of the 17th International Conference on
  Autonomous Agents and MultiAgent Systems (AAMAS)}, pp.\  2085--2087, 2018.

\bibitem[Tan(1993)]{Tan93}
Tan, M.
\newblock Multi-agent reinforcement learning: Independent vs. cooperative
  agents.
\newblock In \emph{In Proceedings of the Tenth International Conference on
  Machine Learning (ICML)}, pp.\  330--337, 1993.

\bibitem[Tang et~al.(2017)Tang, Houthooft, Foote, Stooke, Xi~Chen, Duan,
  Schulman, DeTurck, and Abbeel]{Tang17}
Tang, H., Houthooft, R., Foote, D., Stooke, A., Xi~Chen, O., Duan, Y.,
  Schulman, J., DeTurck, F., and Abbeel, P.
\newblock \#{E}xploration: A study of count-based exploration for deep
  reinforcement learning.
\newblock In \emph{Advances in Neural Information Processing Systems (NIPS)
  30}, pp.\  2753--2762. 2017.

\bibitem[van~den Oord et~al.(2016)van~den Oord, Kalchbrenner, Espeholt,
  kavukcuoglu, Vinyals, and Graves]{vandenOord16}
van~den Oord, A., Kalchbrenner, N., Espeholt, L., kavukcuoglu, k., Vinyals, O.,
  and Graves, A.
\newblock Conditional image generation with {P}ixel{CNN} decoders.
\newblock In \emph{Advances in Neural Information Processing Systems (NIPS)
  29}, pp.\  4790--4798. 2016.

\bibitem[Vinyals et~al.(2019)Vinyals, Babuschkin, Chung, Mathieu, Jaderberg,
  Czarnecki, Dudzik, Huang, Georgiev, ichard Powell, Ewalds, Horgan, Kroiss,
  Danihelka, Agapiou, Oh, Dalibard, Choi, Sifre, Sulsky, Vezhnevets, Molloy,
  Cai, Budden, Paine, Gulcehre, Wang, Pfaff, Pohlen, Wu, Yogatama, Cohen,
  McKinney, Smith, Schaul, Lillicrap, Apps, Kavukcuoglu, Hassabis, and
  Silver]{Vinyals19}
Vinyals, O., Babuschkin, I., Chung, J., Mathieu, M., Jaderberg, M., Czarnecki,
  W., Dudzik, A., Huang, A., Georgiev, P., ichard Powell, Ewalds, T., Horgan,
  D., Kroiss, M., Danihelka, I., Agapiou, J., Oh, J., Dalibard, V., Choi, D.,
  Sifre, L., Sulsky, Y., Vezhnevets, S., Molloy, J., Cai, T., Budden, D.,
  Paine, T., Gulcehre, C., Wang, Z., Pfaff, T., Pohlen, T., Wu, Y., Yogatama,
  D., Cohen, J., McKinney, K., Smith, O., Schaul, T., Lillicrap, T., Apps, C.,
  Kavukcuoglu, K., Hassabis, D., and Silver, D.
\newblock {A}lpha{S}tar: Mastering the real-time strategy game {S}tar{C}raft
  {II}.
\newblock Deepmind blog, accessed 04/16/2019,
  \href{https://deepmind.com/blog/alphastar-mastering-real-time-strategy-game-starcraft-ii}{\texttt{https://deepmind.com/blog/alphastar-ma\\stering-real-time-strategy-game-starcr\\aft-ii}},
  2019.

\bibitem[Watkins \& Dayan(1992)Watkins and Dayan]{Watkins92}
Watkins, C. and Dayan, P.
\newblock Q-learning.
\newblock \emph{Machine Learning}, 8:\penalty0 279--292, 1992.

\bibitem[Watkins(1989)]{Watkins89}
Watkins, C. J. C.~H.
\newblock Learning from delayed rewards.
\newblock PhD thesis, Cambridge University, 1989.

\bibitem[Zheng \& Yue(2018)Zheng and Yue]{Zheng18}
Zheng, S. and Yue, Y.
\newblock Structured exploration via hierarchical variational policy networks,
  2018.
\newblock URL \url{https://openreview.net/forum?id=HyunpgbR-}.

\end{thebibliography}
